# Survey of Pseudonymization, Abstractive Summarization & Spell Checker for Hindi and Marathi

**Rasika Ransing, Mohammed Amaan Dhamaskar, Ayush Rajpurohit, Amey Dhoke, Sanket Dalvi**

Vidyalankar Institute of Technology
{rasika.ransing, mohammedamaan.dhamaskar, ayush.rajpurohit, amey.dhoke, sanket.dalvi}@vit.edu.in

**Abstract**

India's vast linguistic diversity presents unique challenges and opportunities for technological advancement, especially in the realm of Natural Language Processing (NLP). While there has been significant progress in NLP applications for widely spoken languages, the regional languages of India, such as Marathi and Hindi, remain underserved. Research in the field of NLP for Indian regional languages is at a formative stage and holds immense significance. The paper aims to build a platform which enables the user to use various features like text anonymization, abstractive text summarization and spell checking in English, Hindi and Marathi language. The aim of these tools is to serve enterprise and consumer clients who predominantly use Indian Regional Languages.

## 1 Introduction

In the exponential expansion of digital communications and India's progressing toward the "Digital India Campaign", the necessity for tools and services that can fulfil the need of people to understand the texts and documents in Indian regional languages has expanded. While existing text processing tools effectively support prominent languages such as English, the development of comparable services for Indian regional languages remains significantly underdeveloped, hindering progress in these domains.

This platform will be designed to empower services in regional languages like Hindi and Marathi which are two of the most spoken languages in India by leveraging advanced NLP techniques. The platform will incorporate following core functionalities:

1. Text anonymization: In today's highly connected world, there are vast amounts of data that is generated every minute. These large volumes of data must be collected, stored and indexed before it is used for various downstream tasks. However, most of this data is strife with Personally Identifiable Information (PII). PII can be any information like Name, Gender, Organization, etc which can be used independently or in combination with other PII to identify an individual or an organization. The exposure of PII can have severe consequences including identity theft, financial fraud, and reputational damage. The potential for misuse of PII is especially concerning in contexts where sensitive information, such as medical records, financial data, or biometric details, is involved. To solve this issue, all data points should be anonymized before publishing. Anonymization simply means removing this sensitive information. This step is crucial to protect individuals' privacy and mitigate the risks associated with the PII exposure.
2. Text summarization: One of today's major challenges is managing information overload so users can quickly grasp key ideas from large documents. Summaries enhance the efficiency of information sharing, enabling people to understand essential points regardless of their literacy skills. By offering a quick overview, summaries help users decide the relevance of a document. This is particularly valuable in fields like research, legal analysis, and data extraction, where finding the right information is crucial. In NLP, summarization also supports other features like text analysis, translation, and sentiment analysis.

3. Spell correction: With the increase in various tools and platforms, there is a need for people to effectively and correctly present their ideas. For any writing tool having a basic spell checker is essential as it improves accuracy and efficiency of the entire process. For non-native speakers or learners having a good spell checker which supports various languages improves the overall experience of writing content in any desired language. This paper talks about various techniques used for implementing multilingual spell checkers.

The primary objective of this platform is to facilitate communications in these languages to progress towards the digital India future in hand with the regional languages. By fulfilling the demands for text processing in these languages the platform will seamlessly facilitate the services in English along with Hindi and Marathi.

This paper explores the implementation of existing systems and technologies along with approaches and methodologies that will be followed to design the platform. The following sections will detail the literature survey and proposed system along with the potential impact of the platform on multilingual services.

## 2 Literature Survey

### 2.1 Text Anonymization

Anonymization is a data protection technique that involves transforming or removing Personally Identifiable Information (PII) from data to make it unidentifiable. Anonymization has two stages: first is the detection of sensitive information from a given text and second is replacing the detected sensitive information with placeholder or random values. For the first stage, there are various techniques to detect PII like Rule Based tagging, dictionary lookups, part-of-speech (POS) tagging and named entity recognition (NER), but the latter two offer the best results, especially while handling out of vocab terms. POS tagging involves tokenizing sentences into words or tokens and assigning a grammatical class to each token. When POS tagging was introduced, it used classical techniques like Maximum Entropy Modelling [1]; however with the advancements in natural language processing, it evolved to using various modern machine learning models like LSTMs, CRFs and neural networks [2]. While part-of-speech was quite effective during its advent, it was later overshadowed by the advancements in named entity recognition. The term 'named entity' was first introduced at the sixth Message Understanding Conference [3], where the task of such a system was to detect classic entities like persons, locations and organizations in text. Since then, the task of named entity recognition has evolved to also include fine-grained NER to detect subcategories in entities [4] and nested NER to detect nested entities [5].

[6] and [7] compare the use of CRFs, LSTMs, Language Models and Transformer models for Named Entity Recognition (NER). Both of these works highlight the efficiency of transformer-based NER models, which consistently outperform traditional approaches. While these results are remarkable, much like most of the work in NLP, they focus only on the English language.

Across the past works on the Hindi language, FIRE 2014 [8] dataset (NER dataset for English, Hindi, Tamil and Malayalam) and WikiANN [9] dataset (NER dataset of 10,000 sentences for 282 languages) have garnered a lot of interest and led to multiple adaptations of different pretrained models for the task of Hindi NER. However, the first true gold standard dataset for Hindi NER was the HiNER [10] dataset and corresponding NER models released in 2022. It comprises over 100,000 manually annotated sentences and 11 entity categories.

Similarly for Marathi, until recent years, the only available NER corpus were the WikiANN dataset (included Marathi language) and the IIT Bombay Marathi NER Corpus [11], both of which were limited in dataset size and the number of entities. A breakthrough in the field of Marathi NER was the L3Cube-MahaNER [12] dataset and BERT models, which comprises 25,000 manually annotated sentences and 7 entity categories. Another noteworthy mention is the Naamapadam [13] dataset and IndicNER models which provide a NER corpus for 11 Indian languages including Hindi and Marathi. We aim to further expand on this previous work and develop a single model which can perform NER for Hindi and Marathi languages.

The second stage involves the replacement of the detected entities with certain placeholder values. There are various anonymization techniques like Suppression (replaces each character of detected entity with '*'), Permutation (changes sequence of characters), Hashing (generates unique hash for each detected entity), Generalization (replaces detected entity with a generalized term (e.g. - Zip Code is replaced by the city name)) and Pseudonymization (replaces the detected PII with a semantically similar placeholder).

While all these techniques provide privacy and prevent the exposure of PII, Pseudonymization is not only the most sophisticated, but also the most effective in preserving the semantic usefulness of the context. This makes it the most effective technique in sanitizing text and datasets for different NLP tasks. The generation of pseudonyms for replacement is a complex task, and is driven by various characteristics like token length, variation in token styles and sub-categories within entities.

Due to these complexities, pseudonym generation is an active area of research in the NLP space, with various approaches like Dictionary lookups, rule-based matching and the use of generative models. [14] uses Dictionary lookup for Pseudonymization. The dictionary was prepared by curating data points from WikiData and replacements were chosen at random from the dictionary. [15] also compared anonymization across methods involving LLMs to generate pseudonyms and Seq-2Seq model. [16] uses NER for detection and fine-tuned BERT models for pseudonym generation. While these works are quite comprehensive in their research, most of this research is limited to the English language, with little work being done for Indian regional languages. Our research aims to implement these techniques for Hindi and Marathi language and publish a series of Pseudonym generation models for the same.

### 2.2 Abstractive Text Summarization

Early approaches to text summarization relied predominantly on extractive approaches, but with the significant advancement in the field of Natural Language Processing the opportunities to perform the task of summarizing via an advanced or more efficient approach is possible. Abstractive text summarization, a more efficient approach of extractive summarization is possible using the neural network models to generate more coherent and human like summaries.

In [17] the authors investigated several automatic text summarization (ATS) systems, both abstractive and extractive methods. They highlight the importance of artificial intelligence, particularly the use of the sequence-to-sequence (seq2seq) model in extracting of central ideas. These models are different from the extractive techniques in way that they re-word content to express main ideas rather than simply taking out sentences that are important. The journal also examines recurrent neural networks (RNNs), long short-term memory networks (LSTMs), and transformer-based models as the standard architectures used to develop summarization methods. [18] proposes Hindi text summarization with a challenge and introduces IndicBART, a transformer model that is pre-trained on a multilingual corpus and fine-tuned on Hindi. Through transfer learning, IndicBART shares information from the English language to be effective in Hindi, gaining a ROUGE-1 F1 score of 0.544 for abstractive summaries, thus already promising despite little data.

[19] and [20] focus on abstractive summarization for Marathi. Even though transformer-based models improved summary quality significantly, they do not reach human-like results. Paper [19] reports attention-based and stacked LSTM Seq2Seq models using stop-word and rare-word lists for Hindi and Marathi with ROUGE scores of 0.61 (recall) and 0.625 (precision) whereas [20] presents a Deep Belief Network and Decision Tree based approach that achieves 95.49% precision and 92.76% accuracy.

[21] provides a detailed overview of how the script, particularly Hindi, has evolved regarding abstractive text summarization. Like [19], the discussion revolves around the complexity of the script and sentence structure. The authors compare traditional rule-based systems with modern neural networks and explain why these models are effective while processing Hindi text. Detailed inspection of seq2seq models [22] shows that they can be used for Hindi, mainly due to the effectiveness of these models in handling complex sentences and in producing coherent summaries.

[17] rightly emphasizes the need for attention mechanisms for improving the quality of Hindi summaries. In fact, promising results have already been observed with the transformer-based BERT, GPT & T5 fine-tuned models for Hindi, following pre-training on large multilingual corpora.

[23] proposed a transliteration-based approach to improving Hindi text summarization. The authors recommend representing Hindi text in Romanized form before processing. This removes all word segmentation and character ambiguities. The results in fine-tuning with this method generated a 0.5510 ROUGE-1 score. The authors address the potential benefits of this method over classic techniques regarding both semantic coherence and fluency in adaptation. New possibilities of cross-lingual adaptation of NLP models arise with this approach as well.

[24] aims at creating a news summarization tool that includes many ML models, including BERT and TextRank, in combination with NLP techniques like Named Entity Recognition (NER). It is mainly extractive in nature, but this marks the first step towards abstractive techniques by improving the contextual knowledge developed in the machine and generating human-like summaries. The combination of NLP and ML marks the transition from traditional techniques towards advanced neural network-based techniques for making the extraction of the news summarization.

[25] describes in detail datasets for Indian languages, which are significant for training models in regional languages. The authors have pointed out a problem of such scarce annotated datasets for abstractive summarization systems. They offer a comparative analysis of these datasets based on size, diversity, and content. The analysis stands out to be crucial for understanding current progress in such diverse datasets. As stated in [19] and [20], the key role that transfer learning and pre-trained models play is distinguished to overcome possible limitations due to scarcity of large dataset for Indian regional languages.

[26] presents a Hindi text summarization method that is a combination of extractive and abstractive methods. This is done by using a two-step process: first by choosing important sentences through the feature positions, keyword frequency, and title similarity, and then, after removing the redundant contents, making a good summary of the text. The method would have been better if it was rather based on sentence selection and elimination, however, the authors imply that combining modern methods, e.g. neural networks or transformers, will cause a significant improvement.

## 2.3 Spell checker

One of the earliest approaches to spell checking involved dictionary-based techniques. These methods identify misspelled words by checking each word against a predefined dictionary. While this approach works well for standard languages with comprehensive lexicons, it becomes challenging when applied to regional languages such as Marathi and Hindi. These languages are underrepresented in digital corpora, making dictionary creation and maintenance difficult [27].

Another approach includes probabilistic and machine learning-based methods. N-gram models, for example, analyze word sequences to predict the most likely word combinations, which aids in detecting and correcting spelling errors within context. In languages like Hindi and Marathi, where context plays a significant role in meaning, these methods have shown promise. However, challenges remain in obtaining large enough datasets for these models to function effectively in non-English languages [27,28].

Another modern approach involves the use of edit distance algorithms, such as the Levenshtein distance, to quantify how different two strings are and suggest corrections based on the closest matching word. While edit distance methods work well for detecting simple typos, they struggle with more complex errors, particularly in languages with script and orthography distinct from English.

Hybrid methods that combine dictionary-based techniques with machine learning models have also been explored in recent years. These methods aim to leverage the strengths of both approaches by using machine learning to identify likely corrections while still ensuring that suggestions conform to the basic linguistic rules of the language. Such techniques are especially useful in handling challenges posed by the morphological complexity of Marathi and Hindi [27,28].

One major obstacle is the lack of extensive linguistic resources, such as annotated corpora and comprehensive dictionaries, for these languages. Unlike English, where vast amounts of data are available for training and testing, Indian languages suffer from a scarcity of high-quality linguistic datasets. This limits the performance of both rule-based and machine learning-driven spell checkers, as they require large datasets to function effectively [29].

[29] explores the intricacies of creating a spell-checking system for the Shahmukhi script, a less commonly studied script used in the Punjabi language. The design considerations, such as the identification of specific phonetic variations, typographical errors, and the integration of linguistic rules, offer valuable insights that can be extended to other regional languages like Marathi and Hindi. The method presented uses rule-based and statistical approaches to develop a comprehensive spell checker that considers the linguistic characteristics of the target language. This combination of approaches ensures a higher rate of error detection and correction, which is crucial for regional languages with rich morphological structures [30].

In the context of multilingual NLP systems, the application of spell-checking algorithms must address not only the diverse character sets and scripts but also the nuances in grammar and vocabulary. The Shahmukhi spell checker highlights how leveraging specific linguistic features, such as affixes and phonetic patterns, can be pivotal for improving the spell-checking accuracy in non-Latin scripts [29]. The inclusion of a large and well-annotated corpus further enhances the system's performance, a strategy that is equally applicable to Marathi and Hindi. The use of dictionaries and corpora can greatly aid in developing spell checkers that are not only accurate but also contextually aware, reducing the likelihood of inappropriate corrections.

## 3  Proposed System

### 3.1  Text Anonymization

We aim to develop an anonymization engine for Indian Regional languages, using pseudonymization [15] technique, with a focus on Hindi and Marathi. This requires the use of a NER model for identification of PII and using custom techniques to implement pseudonymization. We plan to experiment with various generative models to generate pseudonyms for Hindi and Marathi.

The language specific NER models outlined in [10] and [12] appear to outperform IndicNER, as indicated by their respective evaluation techniques. However, there is a scarcity of such language specific NER models, due to a lack of a large corpus. We aim to address this problem by training a multilingual NER model to handle both Hindi and Marathi language and compare our results with that of MahaNER and HiNER based models.

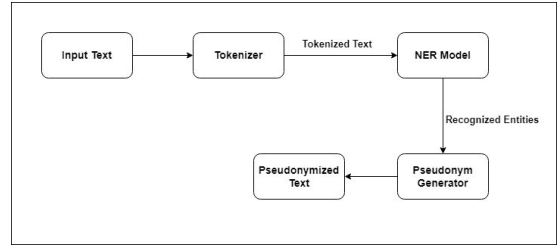

Figure 1: Proposed System for Text Anonymization

### 3.2  Abstractive Text Summarization

The task for abstractive text summarization for Hindi would be carried out using the "*Hindi Text Short and Large Summarization Corpus*" [31,32] dataset along with the Hindi language dataset provided by ILSUM.

For Marathi summarization we chose to translate a dataset provided by "*CNN-DailyMail News Text Summarization*" [33] with number of instances over 300k into Marathi language out of which we have already complete the translation of approximately 60k instances at the time of writing this survey paper.

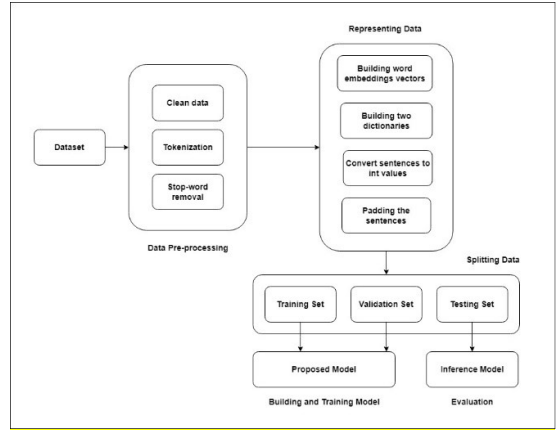

Figure 2 Abstractive Text Summarization Proposed System

The model we chose to use are:

#### 3.2.1  BART based model

**IndicBART** provided by AI4Bharat based on BART which is a denoising autoencoder that maps a corrupted document to the original document through a sequence-to-sequence model with a bidirectional encoder over corrupted text and a left-to-right autoregressive decoder [18].

**Fine-tuning Facebook's BART-large-cnn model** involves adapting the pre-trained model, optimized for English text summarization, to work with Marathi and Hindi datasets allowing benefits of pre-existing fine-tuning on CNN dataset.

### 3.2.2 LSTM-based models

**The Time Distributed Stacked LSTM** [19] model involves stacking two LSTM layers in the encoder making the model effective for capturing complex patterns in text data and hence making it useful for summarizing tasks. The first LSTM layer processes the input sequence, and its output is fed into the second LSTM layer.

The Attention mechanism is a significant enhancement to LSTM models, allowing the network to focus on specific parts of the input sequence. In the **Attention-Based Stacked LSTM** model [19], three LSTM layers are stacked together. This is achieved by using the attention helps the model decide which words in the input sequence are most relevant to the current word being generated in the summary.

### 3.2.3 Deep Recurrent Neural Networks (DRNN)

Deep Recurrent Neural Networks (DRNN) are an extension of traditional RNNs with multiple hidden layers by stacking RNN layers, enabling the network to capture more complex patterns in the data [21] and overcome limitations of shallow RNNs allows model to capture intricate patterns in text.

### 3.3 Spell checker

Our approach for Spell checker includes four main stages: Error Detection, Generating Candidate Suggestion, Ranking Suggestions and Replacing Incorrect word with correct word.

### 3.3.1 Error Detection:

We keep our error-search strictly to non-words errors; for every token in a sentence, we check for its occurrence in the dictionary.

### 3.3.2 Generating Candidate Suggestion:

Given an unknown token, we generated a list of all known words within an edit distance of 2, calling them candidate suggestions. Two intuitive approaches for generating suggestions that work reasonably well on smaller datasets are: checking the edit distance of the misspelled word against all words in the dictionary faces challenge with large corpora, and generating a list of all words within an edit distance of 2 from the incorrect spelling faces challenges with longer words.

### 3.3.3 Ranking Suggestions:

We use Minimum edit distance to rank the suggestions. Minimum edit distance measures how many operations are required to change one word into another. These operations can include:

- **Insertion**: Adding a character.
- **Deletion**: Removing a character.
- **Substitution**: Replacing one character with another.
- **Transposition**: Swapping two adjacent characters.

By calculating the edit distance for all possible correct words, we can rank them based on their distance from the incorrect word.

### 3.3.4 Replacing Incorrect word with Correct word:

At the end we compare the result from n-grams and the top-k result from suggested correct words and if there is a match in those two results, the matched word is replaced with the incorrect word. If there is no match in those two results, the top ranked suggested result is replaced with the incorrect word.

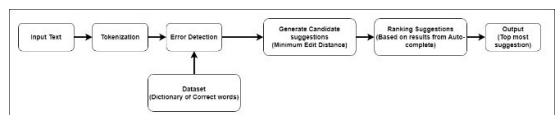

Figure 3: Spell Checker Proposed System

## 4 Conclusion

Recent advancement in the field of Natural Language Processing (NLP) have shown remarkable progress in applications namely text summarization, spell checking and data anonymization. But the development of equivalent tools in Hindi and Marathi is relatively nascent. While some research exists, a unified platform integrating various NLP tasks is still lacking.

The research highlights the potential of developing a user-friendly platform for text summarization, spell checking, anonymization and sentiment analysis especially for Marathi and Hindi language which significantly contribute to digital inclusion. Our proposed system focuses on addressing challenges like data scarcity and language complexity by implementing state-of-the-art AI and ML technologies by combining available resources with our own datasets.